%% file: main.tex
\documentclass[10pt,twocolumn,letterpaper]{article}

\usepackage{cvpr}              %

\input{preamble}

\definecolor{cvprblue}{rgb}{0.21,0.49,0.74}

\usepackage[pagebackref,breaklinks,colorlinks,allcolors=cvprblue]{hyperref}
\usepackage{multirow}
\usepackage{arydshln}
\usepackage{amsmath}
\usepackage{wasysym}
\usepackage{mathtools}
\usepackage[accsupp]{axessibility}

\def\paperID{}
\def\confName{}
\def\confYear{}

\title{Need for Speed: Zero-Shot Depth Completion with Single-Step Diffusion}

\author{
  Jakub Gregorek$^{1,2}$ \quad Paraskevas Pegios$^{1,2}$ \quad Nando Metzger$^{3}$ \quad Konrad Schindler$^{3}$ \\ Theodora Kontogianni$^{1,2}$ \quad Lazaros Nalpantidis$^{1,2}$
\\
\\
  $^{1}$DTU - Technical University of Denmark \qquad $^{2}$Pioneer Centre for AI \qquad  $^{3}$ETH Zürich
}

\begin{document}

\maketitle

\input{0_abstract} 
\input{1_introduction}
\input{2_related_work}
\input{3_method}
\input{4_experimental_setup}

\input{5_results}

\input{6_discussion}
\input{7_conclusion}
\clearpage
{
    \small
    \bibliographystyle{ieeenat_fullname}
    \bibliography{main}
}

\appendix

\end{document}

%% file: preamble.tex
\usepackage[table]{xcolor}

%% file: 0_abstract.tex
\begin{abstract}

We introduce Marigold‑SSD, a single‑step, late‑fusion depth completion framework 
that leverages strong diffusion priors while eliminating the costly test‑time optimization 
typically associated with diffusion‑based methods. By shifting computational burden 
from inference to finetuning, our approach enables efficient and robust 3D perception 
under real‑world latency constraints. Marigold‑SSD achieves significantly faster inference with 
a training cost of only 4.5 GPU days. We evaluate our method across four indoor and 
two outdoor benchmarks, demonstrating strong cross‑domain generalization and 
zero‑shot performance compared to existing depth completion approaches.
Our approach significantly narrows the efficiency gap between diffusion‑based and 
discriminative models. Finally, we challenge common evaluation protocols by 
analyzing performance under varying input sparsity levels.
Page: \url{https://dtu-pas.github.io/marigold-ssd/}

\end{abstract}

%% file: 1_introduction.tex
\section{Introduction}

\input{figures/compare-speed/compare-speed}

Depth completion aims to recover a dense depth map from sparse measurements given an input RGB image and is a core task for applications such as autonomous driving,  robotics, and 3D reconstruction~\cite{depth-completion-survey, depth-completion-recent-advances}. In real-world settings, depth sensors such as LiDAR  provide only partial information, while downstream tasks require dense depth maps to  reason about scene structure. Despite significant progress, many existing methods rely on discriminative models~\cite{nlspn,bp-net,lrru,completion-former,sparsity-agnostic-dc} whose performance often degrades under varying sparsity patterns and domain shifts, limiting their applicability in open-world scenarios. This has motivated growing interest in zero-shot evaluations and  approaches~\cite{vpp4dc,ogni-dc,marigold-dc,prompt-da},  where models are expected to generalize without dataset-specific retraining.

\input{figures/compare-architectures/compare-architectures}
Recent work~\cite{marigold,depth-anything,depth-pro,depth-fm,geowizard} leverages strong visual priors learned by foundation models trained on large-scale data~\cite{stable-diffusion,dinov2}. Among these, generative diffusion-based methods~\cite{he2024lotus, depth-fm,krishnan2025orchid,geowizard}, such as Marigold~\cite{marigold}, which repurposed Stable Diffusion~\cite{stable-diffusion} for depth estimation, have proven particularly effective by encoding rich semantic and geometric structure through iterative denoising. Recently, Marigold has been extended trough test-time optimization to depth completion~\cite{marigold-dc} and depth inpainting~\cite{steered-marigold}, consistently outperforming discriminative approaches in zero-shot settings. However, performance comes at a significant computational cost, as inference typically requires tens or hundreds of denoising steps and previous methods often require ensembling strategies~\cite{marigold-dc}, making them impractical for embodied AI applications.

In this work, we focus on reducing the computational complexity of diffusion-based methods for zero-shot depth completion. We argue that iterative paradigms are not strictly necessary to achieve high-quality results. Building on top of \textbf{Marigold}~\cite{marigold} and its variants~\cite{marigold-dc,steered-marigold,marigold-e2e},
we propose zero-shot depth completion with \textbf{S}ingle-\textbf{S}tep \textbf{D}iffusion (\textbf{Marigold-SSD}), a diffusion-based method that occupies a unique region of the performance-speed trade-off space. As illustrated in Fig.~\ref{fig:compare-speed}, Marigold-SSD achieves performance comparable to state-of-the-art iterative methods, while being orders of magnitude faster, significantly narrowing the gap between slow but robust diffusion-based approaches and fast discriminative methods.

We demonstrate a single-step diffusion framework for zero-shot depth completion with end-to-end fine-tuning. To enable conditioning on sparse measurements we introduce a late-fusion conditional decoder that injects the condition during decoding. Our design leverages the diffusion prior and enables single-step inference after fine-tuning which takes approximately only 4.5 days on a single NVIDIA H100 GPU. By shifting computation from inference to finetuning we provide a practical path towards narrowing the gap between performance and runtime efficiency. A comparison with test-time optimization approaches is shown in Fig.~\ref{fig:compare-architectures}. Our method yields strong zero-shot performance across four diverse indoor and two outdoor benchmarks. 

Our main contributions are the following:
\begin{itemize}
    \item The first single-step diffusion-based method for depth completion, significantly faster than diffusion baselines while delivering better performance on average, and remaining competitive even when baselines employ ensembling at substantially higher computational cost.
    \item A simple yet effective late-fusion strategy for conditioning on sparse measurements, whose effectiveness against early-fusion is validated through ablation studies.
    \item A comprehensive zero-shot evaluation across indoor and outdoor datasets, demonstrating strong robustness of Marigold-SSD to varying condition sparsity levels, while revealing limitations of existing evaluation benchmarks.
\end{itemize}

%% file: figures/compare-speed/compare-speed.tex
\begin{figure}[t]
    \centering
    \includegraphics[width=1\linewidth]{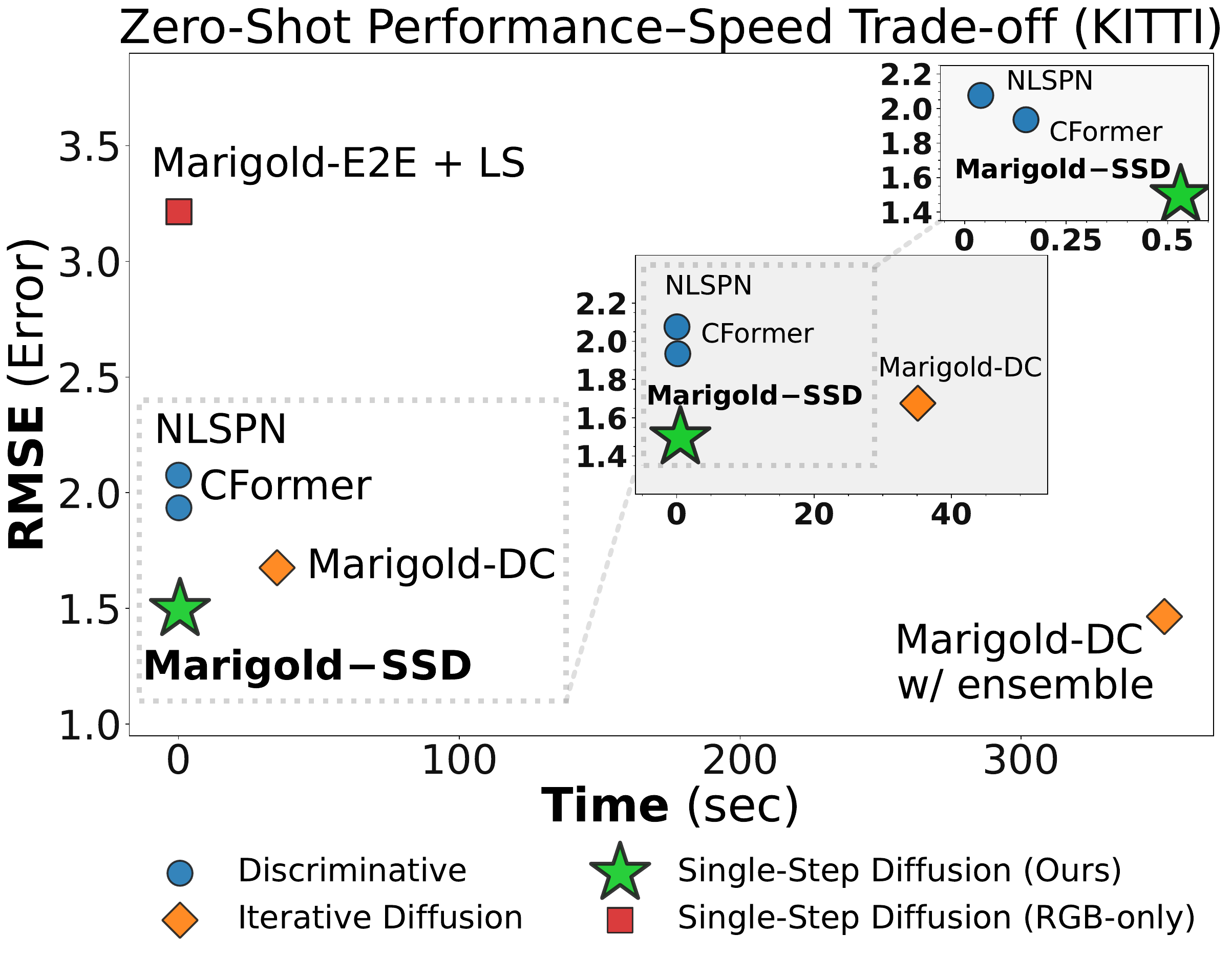}
    \caption{\textbf{Performance vs speed trade-off.} Comparison of our method \textbf{Marigold-SSD} with other diffusion-based approaches Marigold-DC~\cite{marigold-dc} and Marigold-E2E~\cite{marigold-e2e} + LS (w/o sparse condition) as well as discriminative baselines~\cite{nlspn,completion-former} on KITTI dataset~\cite{kitti-dataset}. \textbf{Marigold-SSD} occupies a unique region in the trade-off space closing the efficiency gap to discriminative methods while retaining the benefit of the strong diffusion prior.
    }
    \label{fig:compare-speed}
\end{figure}

%% file: figures/compare-architectures/compare-architectures.tex
\begin{figure*}[t]
    \begin{subfigure}{.49\linewidth}
        \centering
        \includegraphics[width=1.0\linewidth]{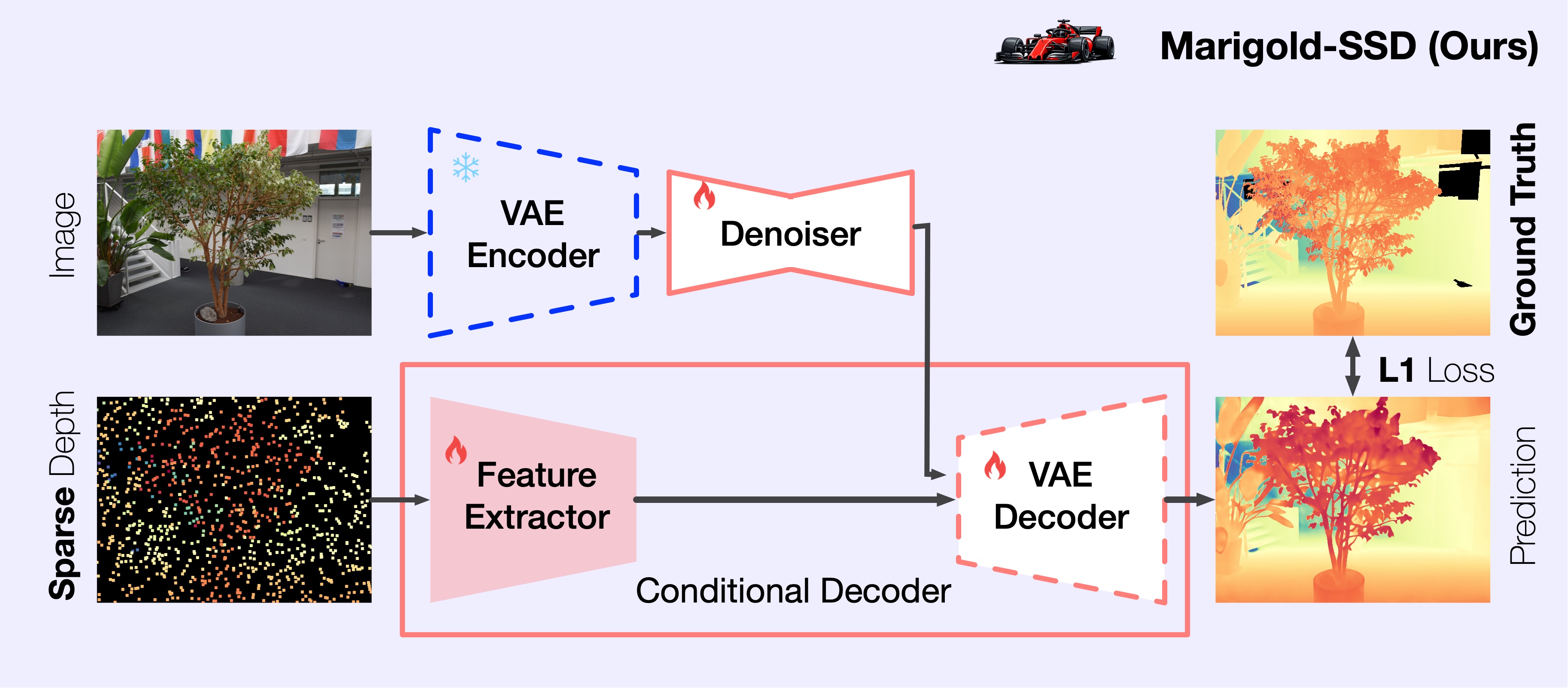}
    \end{subfigure}
    \begin{subfigure}{.49\linewidth}
        \centering
        \includegraphics[width=1.0\linewidth]{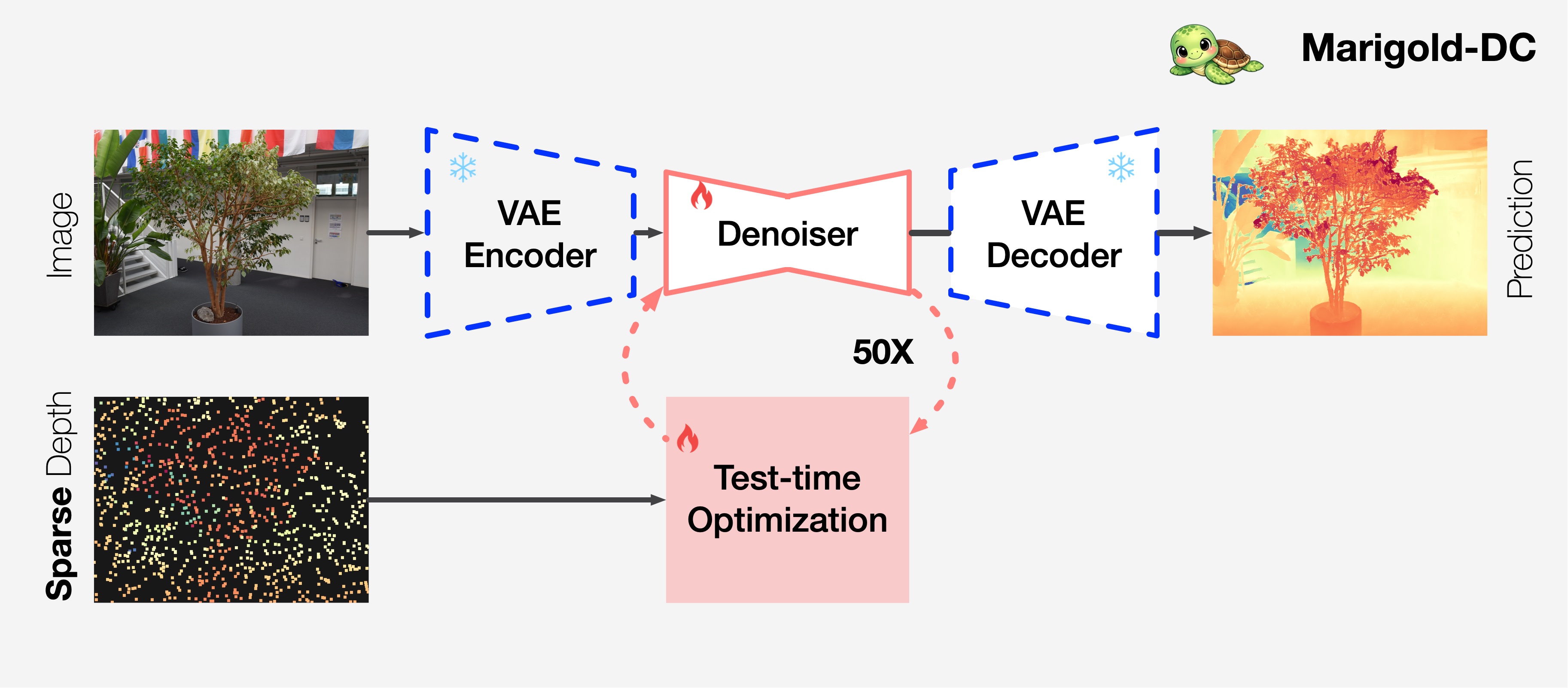}
    \end{subfigure}
	\caption{
        \textbf{Marigold-SSD for zero-shot depth completion}. We present a single-step diffusion framework with end-to-end fine-tuning as an efficient alternative to the test-time optimization approach of Marigold-DC~\cite{marigold-dc}. To this end, we introduce a conditional decoder with late fusion to incorporate sparse depth measurements. At inference, our method Marigold-SSD produces high-quality results in a single step, while Marigold-DC typically requires 50 optimization steps per inference and often ensembling 10 inferences for further improvements.
    }
    \label{fig:compare-architectures}
\end{figure*}

%% file: 2_related_work.tex
\section{Related Work}

\input{figures/conditional-decoder/conditional-decoder}
\noindent \textbf{Zero-Shot Depth Estimation.}
The zero-shot depth estimation was spearheaded by MiDaS \cite{midas}, introducing 
parametrization and losses allowing mixing training datasets originating from different
sources, including 3D movies. The follow up work ZoeDepth \cite{zoe-depth} innovated
on the architecture, and introduced metric bins allowing to estimate metric depth. 
Piccinelli et al.~\cite{uni-depth,unidepth2} proposed methods disentangling camera 
parameters from geometry of the 3D scene exploiting spherical representation.
DepthAnything \cite{depth-anything} takes advantage of large amount of diverse and 
unlabeled data. The subsequent work benefited from replacing real data by synthetic
and scaling up the teacher model~\cite{depth-anything-v2}. Another discriminative 
model, Depth Pro \cite{depth-pro}, was focusing on boundary accuracy and cross-domain 
focal-length estimation. Wang et al.~\cite{moge} takes advantage of geometry supervision 
techniques, like point cloud alignment solver and multi-scale alignment loss. The work
was later extended for metric detph estimation~\cite{moge2}. Majority of the
latest models~\cite{uni-depth,unidepth2,depth-anything,depth-anything-v2,
depth-pro,moge,moge2,depth-anything-v3} benefit from the ViT architecture initialized from 
DINOv2~\cite{dinov2}. Even though these models are not grounding their results 
on real measurements, zero-shot depth estimation models are relevant as strong 
priors which can be exploited for the task of depth completion.\\

\noindent \textbf{Depth Completion.}
Spatial Propagation Networks (SPNs) constitute a widely adopted family of 
models for depth completion, originally proposed by Liu et al.~\cite{spn} 
and subsequently extended to more variants~\cite{cspn, cspnpp, nlspn, dyspn}.
SPN modules have further been enhanced in numerous follow‑up 
methods~\cite{penet, semattnet, completion-former, bevdc, nddepth, tri-perspective,
bp-net, depthor, improving-dc, rignetpp}. Most depth completion pipelines leverage 
guidance from a monocular RGB image~\cite{lp-net, decomposition-a-b, guidenet}, 
including the majority of SPN-based approaches. Additional modalities such as 
surface normals~\cite{nddepth} and semantic segmentation~\cite{semattnet} 
have also been explored, alongside completion without explicit guidance~\cite{lidiff}.
While many methods process sparse depth samples projected onto the image plane, 
others employ multi-planar projections~\cite{tri-perspective} or operate directly on 
raw point clouds~\cite{point-dc, 3d-depth-net, gac-net, bevdc}. Wang et al.~\cite{lrru} 
demonstrated the advantages of depth pre‑completion performed via classical image 
processing techniques~\cite{ipbasic}. Several works investigate robustness to varying 
sparsity levels or sampling patterns~\cite{ogni-dc, sparsity-agnostic-dc, steered-marigold, 
vpp4dc, arapis}, and others explicitly address noisy depth measurements~\cite{omni-dc, depthor}.
A large portion of prior work focuses on single-dataset training and does not emphasize 
out‑of‑domain generalization. For zero‑shot scenarios, depth completion can benefit 
from the strong priors of monocular depth estimators~\cite{prompt-da, depthor, depthorpp, 
DMD3C} or the generalization ability of stereo matching networks~\cite{vpp4dc}. 
Approaches vary from direct fine‑tuning~\cite{prompt-da, depthor, depthorpp} 
and distillation~\cite{DMD3C} to test‑time optimization~\cite{test-prompt-dc}.
\\

\noindent \textbf{Diffusion-Based Approaches.}
Diffusion‑based approaches have demonstrated strong performance in zero‑shot depth 
estimation~\cite{marigold,marigold-e2e,prime-depth,pixel-perfect}, depth completion~\cite{marigold-dc}, 
and inpainting~\cite{steered-marigold}. PrimeDepth~\cite{prime-depth} leverages 
Stable~Diffusion~\cite{stable-diffusion} as a feature extractor, while Ke et al. 
fine‑tunes Stable Diffusion~2~\cite{stable-diffusion} for depth estimation 
introducing Marigold~\cite{marigold}. Hybrid methods combining diffusion and discriminative 
models have also been explored~\cite{better-depth,sharp-depth}. The iterative multi‑step nature 
of diffusion enables plug‑and‑play conditioning with sparse depth maps~\cite{marigold-dc,steered-marigold} 
for depth completion. However, this iterative process comes with substantial computational cost. 
A practical path toward broader adoption in resource‑constrained scenarios is reducing the
number of diffusion steps. Ke et al.~\cite{marigold-journal} address this by distilling 
Marigold into a Latent Consistency Model~\cite{lcm} for few‑step inference. Gui et al.~\cite{depth-fm} 
and Xu et al.~\cite{pixel-perfect} re-frame the problem using flow matching~\cite{flow-matching}.
Garcia et al.~\cite{marigold-e2e} fine‑tunes Marigold for single‑step diffusion‑based depth estimation, 
dramatically decreasing inference time. In this work, we bring the single‑step inference to depth completion.

%% file: figures/conditional-decoder/conditional-decoder.tex
\begin{figure*}[t]
\smallskip
\centering
\includegraphics[width=0.86\linewidth]{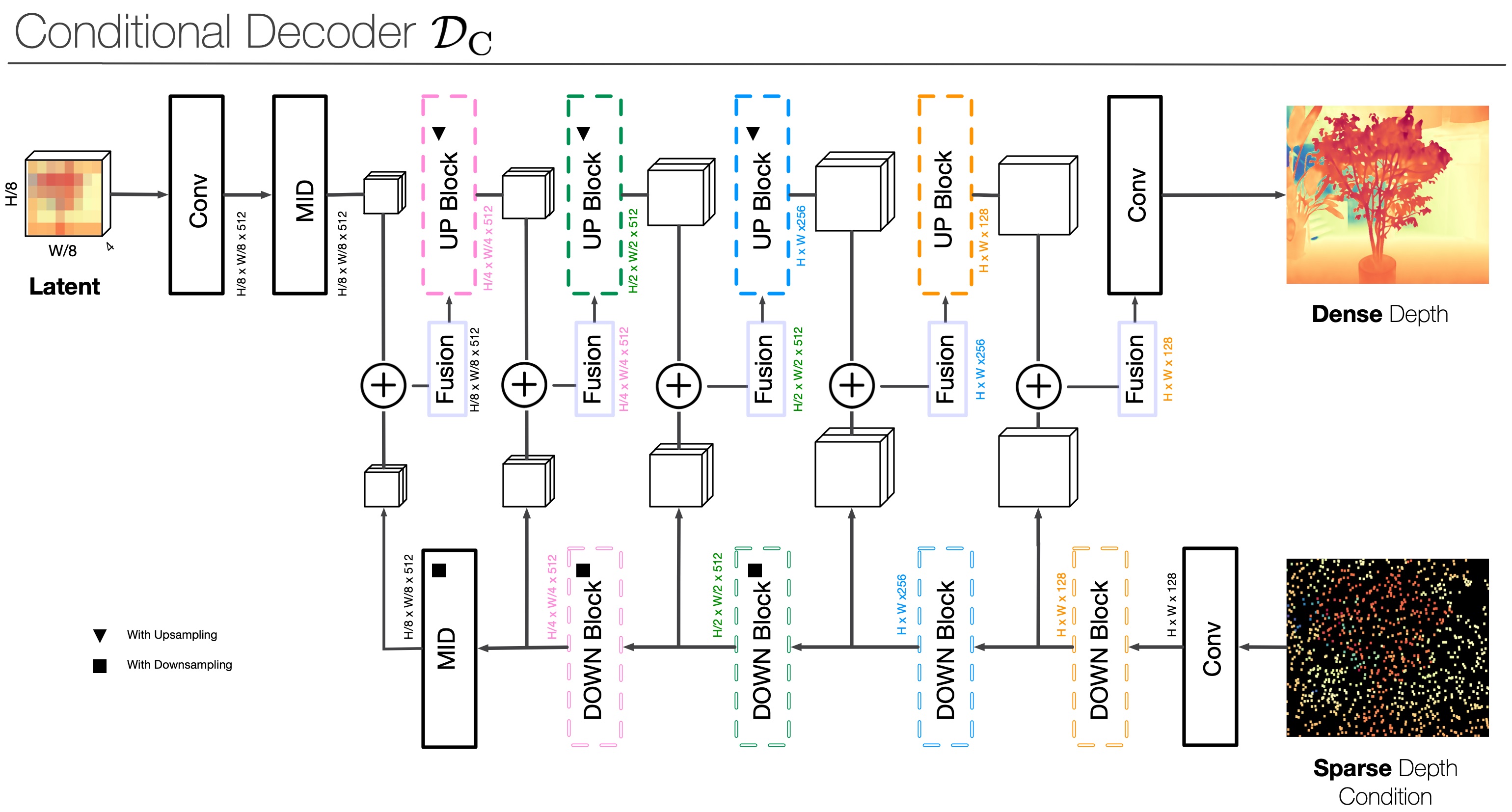}
\caption{
    \textbf{Internal architecture of the conditional decoder.}
    $\mathcal{D}_{\mathbf{C}}$ consists of the VAE decoder $\mathcal{D}$ (top row) and blocks processing the sparse condition~$\mathbf{C}$ (bottom row), adapted from the VAE encoder $\mathcal{E}$ (differing in down-sampling positions). Feature maps are concatenated channel‑wise ($\oplus$) at five levels and the fusion blocks use $1\times1$ convolutions (Eq.~\ref{eq:fusion}). Conv denotes standard convolution layers, UP, DOWN, and MID blocks are ResNet~\cite{resnet}-based, and MID blocks additionally containing an attention layer.
} 
\label{fig:conditional-decoder}
\end{figure*}

%% file: 3_method.tex
\section{Method}

Given an input RGB image $M \in \mathbb{R}^{H \times W \times 3}$ and a sparse depth condition $C \in \mathbb{R}^{H \times W}$, our goal is to predict a dense depth map $D \in \mathbb{R}^{H \times W}$. Our method builds on the generative prior of Marigold~\cite{marigold}, adopts a single-step diffusion formulation~\cite{marigold-e2e}, and extends it to depth completion by incorporating sparse measurements through a late-fusion strategy and an end-to-end fine-tuning scheme.\\

\subsection{Marigold for Diffusion-based Depth Estimation}

  Marigold~\cite{marigold} formulates monocular depth estimation as an conditional diffusion process~\cite{ddpm} in the latent space of a frozen VAE~\cite{van2017neural} with encoder $\mathcal{E}$ and decoder $\mathcal{D}$. The forward process corrupts the clean depth latent $x_0=\mathcal{E}(D)$ by adding Gaussian noise $\epsilon\sim\mathcal{N}(0,I)$ under a variance schedule $\{\beta_t\}_{t=1}^{T}$, where $D$ is normalized within [-1,1] and replicated to three channels to match the VAE input. For a timestep $t\in\{1,\dots,T\}$, the noisy latent is $x_t=\sqrt{\bar{\alpha}_t}x_0+\sqrt{1-\bar{\alpha}_t}\epsilon$, where $\alpha_t=1-\beta_t$ and $\bar{\alpha}_t=\prod_{s=1}^{t}\alpha_s$. Following Stable Diffusion~\cite{stable-diffusion}, Marigold adopts the $v$-parameterization~\cite{salimansprogressive} and is trained with a mean-squared error objective. In particular, a UNet~\cite{ronneberger2015u} denoiser is conditioned on the timestep $t$, the noisy latent $x_t$, and the RGB input encoded into the same latent space as $m=\mathcal{E}(M)$, i.e., $\hat{v}_{t}=v_{\theta}(x_t \oplus m,t)$, where $\oplus$ denotes channel-wise concatenation, and it is trained to regress the target velocity $v_t^{*}=\sqrt{\bar{\alpha}_t}\epsilon-\sqrt{1-\bar{\alpha}_t}x_0$. At inference, Marigold starts from a Gaussian latent $x_T\sim\mathcal{N}(0,I)$ and iteratively applies DDIM~\cite{songdenoising} sampler for $50$ denoising steps. The final latent is decoded to obtain the dense depth prediction, $\hat{D}=\mathcal{D}(\hat{x}_0)$. In practice, Marigold uses test-time ensembling by running inference multiple times, aligning each output with a per-prediction scale and shift, and taking the pixel-wise median of the aligned predictions~\cite{marigold}.

\subsection{Depth Completion with Single-Step Diffusion}

Previous methods such as Marigold-DC~\cite{marigold-dc} leverage Marigold as a strong prior and employ a guided diffusion~\cite{dhariwal2021diffusion,chungdiffusion,weng2024fast} variant that uses sparse conditions for test-time optimization of the depth latent during denoising. Although this achieves strong zero-shot performance, it is expensive, typically requiring 
$50$ steps per inference and ensembling $10$ inferences to improve results. In contrast, as shown in Fig.~\ref{fig:compare-architectures}, we shift computation to fine-tuning, enabling single-step inference.

Garcia et al.~\cite{marigold-e2e} observed that the previously poor single-step behavior of Marigold~\cite{marigold} largely stemmed from an inference scheduler issue that paired timesteps with inconsistent noise levels. Correcting this with a trailing setting~\cite{lin2024common} restored more reliable single-step approximations, which can then be distilled through end-to-end fine-tuning. Thus, we fix the timestep to $t=T$ and set noise to zero. Since $\bar{\alpha}_T \approx 0$, the input $x_T$ contains almost no signal from $x_0$ and we can effectively tune for single-step prediction. Given $m=\mathcal{E}(M)$, the output of the denoiser is used to get the clean latent as $\hat{x}_0=\sqrt{\bar{\alpha}_t}x_t-\sqrt{1-\bar{\alpha}_t}\hat{v}_t$. To adapt the architecture for depth completion: (i) we introduce a conditional decoder $\mathcal{D}_{C,\phi}(\cdot)$ to replace the original VAE decoder and inject sparse measurements $C$, and (ii) we fine-tune the resulting model with a task-specific loss.\\

\noindent \textbf{Late Fusion with Conditional Decoder.} The architecture is illustrated in Fig.~\ref{fig:conditional-decoder}. The conditional decoder takes the predicted depth latent $\hat{x}_0$ and the sparse depth condition $C$, normalized to the same $[-1,1]$ range as $D$, and estimates a dense depth map $\hat{D}$. To inject $C$ in a late-fusion manner, we mirror the multi-scale structure of the original frozen VAE ($\mathcal{E}, \mathcal{D}$). We introduce a trainable condition feature extractor $\mathcal{F}$ that processes $C$ and extracts $L{=}5$ multi-scale feature maps $\{f_l^{\mathcal{F}}\}_{l=1}^{L}$ that match the spatial resolutions of the decoder features $\{f_l^{\mathcal{D_{C}}}\}_{l=1}^{L}$ computed from $\hat{x}_0$. At each level $l$, the high-level sparse conditioning features are fused to together with the dense depth features via convolution layers:
\begin{equation}
    f_l=\texttt{CONV} \left(f_l^{\mathcal{D}}\oplus f_l^{\mathcal{E}}\right),
    \label{eq:fusion}
\end{equation}
where \texttt{CONV} is a $1{\times}1$ convolution. We initialize the decoder from the original frozen VAE decoder $\mathcal{D}$, while the feature extractor is initialized from $\mathcal{E}$. Inspired by ControlNet~\cite{zhang2023adding}, we initialize \texttt{CONV} as a zero convolution layers. This makes the conditioning path output zero at initialization and preserves the behavior of the original VAE decoder. When fine-tuning, the weights gradually increase the contribution of the condition $C$. \\

\noindent \textbf{End-to-End Fine-tuning for Depth Completion.} We fine-tune our model end-to-end with a task loss rather than the diffusion training objective. Previous works in monocular depth estimation~\cite{marigold-e2e} use an affine-invariant loss~\cite{midas}. Instead, we optimize an L1 loss to match depth prediction $\hat{D}$ with the dense target $D$ encouraging consistency with the conditioning sparse measurements $C$. During fine-tuning, we uniformly sample the density of the condition in the range $[\textit{l}\%,\,\textit{h}\%]$, where \textit{l} and \textit{h} are the lower and upper bounds. We initialize our model from Marigold-E2E~\cite{marigold-e2e}, keeping the VAE encoder $\mathcal{E}$ fixed and we fine-tune the proposed conditional decoder $\mathcal{D}_{C, \phi}$ together with the denoising UNet, placing stronger emphasis on our decoder to encourage adaptation for depth completion.  Our design retains the strong diffusion depth prior while enabling high quality predictions during inference with a single-step. \\

\noindent \textbf{Single-Step Inference.} At test time,  we set $x_T$ with zeros for deterministic inference and removing the need for test-time ensembling. Given an RGB input $M$, we compute $m=\mathcal{E}(M)$ and use the denoiser predict the depth latent $\hat{x}_0$. The conditional decoder then combines $\hat{x}_0$ with the sparse condition $C$ to produce $\hat{D}=\mathcal{D}_{C,\phi}(\hat{x}_0,C)$. Since $C$ and the predicted \emph{relative depth} $\hat{D}$ lie in the VAE range, we recover \emph{metric depth} via a global scale and shift $D^{\text{*}}=a\,\hat{D}+b$. The parameters $(a,b)$ are obtained by least-squares alignment to \emph{metric sparse measurements} $C^{\text{*}}$ over valid pixels:
\begin{equation}
\arg\min_{a,b}\sum_{i\in\Omega}\left(a\,\hat{D}_{i}+b-C^{\text{*}}_{i}\right)^{2},
\end{equation}
where $\Omega$ denotes the set of valid sparse depth locations.

%% file: 4_experimental_setup.tex
\section{Experiments and Results}
\subsection{Datasets and Implementation Details}
\input{tables/timing}

\textbf{Training Datasets.} We train our model on the Hypersim~\cite{hypersim} and Virtual~KITTI~\cite{virtual-kitti, virtual-kitti2} synthetic datasets.  Hypersim consists of $461$ indoor scenes; out of which $365$ are  used in the training, totaling in 54K samples. Samples were downscaled to $640\times480$ pixels. Virtual KITTI is a synthetic dataset from the autonomous driving domain.  All 5 scenes of the dataset were used for training in a variety of weather versions  (morning, fog, rain, sunset, and overcast), which is more than 21K samples. For training,  the images were bottom-center cropped to $1216\times352$ pixels.\\

\noindent \textbf{Evaluation Datasets.}
We evaluated the method on \underline{4 indoor datasets}: NYUv2~\cite{nyuv2-dataset},
ScanNet~\cite{scannet-dataset}, VOID~\cite{void-dataset}, and IBims-1~\cite{ibims-dataset}.
NYUv2 and ScanNet were captured with RGB-D sensors, VOID with an active 
stereo setup, and IBims-1 with a laser scanner. 
All indoor datasets were processed at $640\times480$ resolution; NYUv2 results were 
downscaled and cropped to the standard $304\times228$ evaluation size. We used 654 
images from the NYUv2 test split, 745 images from the ScanNet selected by~\cite{marigold-dc}, 
all 800 VOID images, and all 100 IBims-1 images. We match the depth sampling protocol 
of~\cite{marigold-dc}. 500 points were sampled for NYUv2 and ScanNet, 1000 for IBims-1, 
and 1500 points provided by VOID. We also evaluated on \underline{2 outdoor datasets}: 
KITTI~\cite{kitti-dataset, sparsity-invariant-cnns} and DDAD~\cite{ddad-dataset}, 
both using LiDAR and originating from autonomous driving. For KITTI, we cropped the images 
to $1216\times352$ and used the 1000-image validation split. DDAD was processed at $768\times480$ resolution and completed depth
was up-sampled back to full resolution of $1936\times1216$ for evaluation. We used the official validation 
set of 3950 images. Following~\cite{vpp4dc, marigold-dc}, point clouds for both datasets 
were filtered for outliers~\cite{9981654}. On DDAD, approximately 20\% of LiDAR points 
were sampled for guidance, as in~\cite{vpp4dc, ogni-dc, marigold-dc}.\\

\input{tables/timing-kitti}
\noindent \textbf{Implementation Details.} Our implementation is based on HuggingFace diffusers 
library\,\cite{huggingface-diffusers} initializing weights from \,\cite{huggingface-marigold-e2e}.
The training strategy follows\,\cite{marigold-e2e}: mixing Hypersim\,\cite{hypersim} and 
Virtual~KITTI\,\cite{virtual-kitti2} datasets with 9:1 ratio, training for 20K iterations 
utilizing AdamW optimizer\,\cite{adamw}, initial learning rate set to $3 \times 10^{-5}$ for 
$\mathcal{D}_C$ and $3 \times 10^{-6}$ for the UNet, deploying exponential decay 
after 100-step warm-up, accumulating gradients over 32 steps of size 1. We train two models,
sampling the density of depth condition in the ranges $[0.16\%, 5\%]$ and $[0.16\%, 0.5\%]$ 
(models denoted by \APLstar). The first interval covers densities of all evaluation
datasets while the second interval covers density of indoor datasets only. The training was 
performed on a single NVIDIA H100 GPU requiring only 4.5 days per model.\\
\input{tables/quantitative-results3}
\input{figures/qualitative/qualitative}
\input{figures/qualitative/qualitative-kitti}
\input{figures/sparsity-levels/sparsity-levels}

\input{figures/sparsity-levels/sparsity-levels-ddad}

\noindent \textbf{Evaluation Metrics.} Following the common practices for depth completion, 
we report mean absolute error ${MAE} = \frac{1}{N} \sum_{i} \lvert \mathbf{d}_i - \mathbf{g}_i \rvert$, 
and root mean squared error $RMSE = \sqrt{\frac{1}{N} \sum_{i} \lvert \mathbf{d}_i - \mathbf{g}_i \rvert^{2}}$
where $\mathbf{d}_i$ and $\mathbf{g}_i$ denote elements of depth prediction and ground-truth,
and $i \in \{1, ..., N\}$. All reported results are in meters.\\

%% file: tables/timing.tex
\begin{table}[t]
\caption{\textbf{Runtime analysis.}
Average per-image inference time in \textbf{seconds}, throughput in \textbf{FPS}, and \textcolor{green!60!black}{relative speedup} compared to Marigold-DC.
Runtimes are measured on an NVIDIA RTX 4090 GPU. Resolutions shown next to dataset names denote the input size used for the timing experiment. 
Marigold-DC is timed \textbf{without ensembling} (single run); ensembling (e.g., 10 predictions) would increase runtime approximately linearly.
}
\label{tab:timing}
\centering
\setlength{\tabcolsep}{4pt}
\renewcommand{\arraystretch}{1.0}

\resizebox{0.475\textwidth}{!}{%
\begin{tabular}{c l c c c c}
\toprule
& \textbf{Dataset} & \multicolumn{2}{c}{\textbf{Marigold-DC}} & \multicolumn{2}{c}{\textbf{Marigold-SSD} (Ours)} \\
\cmidrule(lr){3-4}\cmidrule(lr){5-6}
&  & \textbf{Time} (s) & \textbf{FPS} & \textbf{Time} (s) ($\uparrow \times$) & \textbf{FPS} \\
\midrule

\multirow{4}{*}{\rotatebox{90}{\tiny \textsc{Indoor}}}
& ScanNet {\tiny(640$\times$480)}
& 25.08 & 0.04
& 0.38 \,\textcolor{green!60!black}{\scriptsize(\textbf{66$\times$})}
& 2.6 \\

& IBims-1 {\tiny(640$\times$480)}
& 24.62 & 0.04
& 0.38 \,\textcolor{green!60!black}{\scriptsize(\textbf{65$\times$})}
& 2.6 \\

& VOID {\tiny(640$\times$480)}
& 24.98 & 0.04
& 0.38 \,\textcolor{green!60!black}{\scriptsize(\textbf{66$\times$})}
& 2.6 \\

& NYUv2 {\tiny(640$\times$480)}
& 24.56 & 0.04
& 0.38 \,\textcolor{green!60!black}{\scriptsize(\textbf{65$\times$})}
& 2.6 \\
\midrule

\multirow{2}{*}{\rotatebox{90}{\tiny \textsc{Outdoor}}}
& KITTI {\tiny(1216$\times$352)}
& 35.10 & 0.03
& 0.53 \,\textcolor{green!60!black}{\scriptsize(\textbf{66$\times$})}
& 1.9 \\

& DDAD {\tiny(768$\times$480)}
& 30.60 & 0.03
& 0.45 \,\textcolor{green!60!black}{\scriptsize(\textbf{68$\times$})}
& 2.2 \\
\midrule

& \textsc{Average}
& \cellcolor[gray]{.9} \textbf{27.49} & \cellcolor[gray]{.9} \textbf{0.04}
& \cellcolor[gray]{.9} \textbf{0.42} \,\textcolor{green!60!black}{\scriptsize(\textbf{66$\times$})}
& \cellcolor[gray]{.9} \textbf{2.4} \\
\bottomrule
\end{tabular}%
}
\end{table}

%% file: tables/timing-kitti.tex
\begin{table}[t]
\caption{\textbf{Speed-performance trade-off.} \emph{Zero-shot} performance on KITTI~\cite{kitti-dataset}. All runtimes are evalauted on NVIDIA RTX 4090 GPUs.  We time Marigold-SSD and Marigold-DC~\cite{marigold-dc} (Table~\ref{tab:timing}) and VPP4DC~\cite{vpp4dc}. Performance is taken from~\cite{marigold-dc}, and runtimes at original resolution for the other discriminative methods from~\cite{tang2026gaussian}.}
\label{tab:timing-kitti}
\centering
\resizebox{0.475\textwidth}{!}{%
\begin{tabular}{l l c c}
\toprule
\textbf{Type} & \textbf{Method} & \textbf{Time} (s) $\downarrow$ & \textbf{RMSE} $\downarrow$ \\
\midrule

\multirow{6}{*}{\textsc{Discriminative}}

& NLSPN~\cite{nlspn}   \textcolor{gray}{\scriptsize (ECCV '20)}   & \; 0.039 & 2.076  \\
& BP-Net~\cite{bp-net}  \textcolor{gray}{\scriptsize (CVPR '24)}   & \; 0.078 & -  \\
& CFormer~\cite{completion-former}  \textcolor{gray}{\scriptsize (CVPR '24)}  & \; 0.151 & 1.935\\
& OGNI-DC~\cite{ogni-dc}  \textcolor{gray}{\scriptsize (ECCV '24)}  & \; 0.266 & - \\
& VPP4DC~\cite{vpp4dc}  \textcolor{gray}{\scriptsize (3DV '24)} & \; 0.164 & 1.609  \\
& GBPN~\cite{tang2026gaussian} \textcolor{gray}{\scriptsize (arXiv preprint '26)}      & \; 0.137 & - \\

\midrule

\multirow{2}{*}{\textsc{Diffusion}}
& \cellcolor[gray]{.9} Marigold-DC~\cite{marigold-dc} \textcolor{gray}{\scriptsize (ICCV '25)}  & \cellcolor[gray]{.9} 35.103 & \cellcolor[gray]{.9} 1.676 \\
& \cellcolor[gray]{.9} Marigold-SSD \textcolor{gray}{\scriptsize (Ours)} & \cellcolor[gray]{.9} \; \textbf{0.527} & \cellcolor[gray]{.9} \textbf{1.496} \\

\bottomrule
\end{tabular}
}
\end{table}

%% file: tables/quantitative-results3.tex
\begin{table*}
\caption{\textbf{Comparison to state-of-the-art on six zero-shot benchmarks.}
Most values are taken from~\cite{marigold-dc}, except for DMD$^{3}$C~\cite{liang2025distilling} and GBPN~\cite{tang2026gaussian}, which are reported from their original papers. Following~\cite{marigold-dc}, we omit BP-Net~\cite{bp-net} and OGNI-DC~\cite{ogni-dc} on NYUv2~\cite{nyuv2-dataset} and KITTI~\cite{kitti-dataset} and SpAgNet~\cite{sparsity-agnostic-dc} on ScanNet~\cite{scannet-dataset} and IBims-1~\cite{ibims-dataset}. Similar to Marigold~\cite{marigold} + optim and Marigold~\cite{marigold} + LS reported in~\cite{marigold-dc}, we evaluate Marigold-E2E~\cite{marigold-e2e} + LS as a single-step diffusion baseline. 
Given \emph{the need for speed}, we highlight \textbf{best} and \underline{second best} excluding the ensemble approach of Marigold-DC\,\cite{marigold-dc}  which increases runtime by an order of magnitude. Our model version trained on lower density-levels is denoted by \APLstar. The rank expresses an average position of the method in the table per metric and dataset.}
\label{tab:quantitative-results}
\centering
\resizebox{\textwidth}{!}{%
\begin{tabular}{cl|cccccccc|cccc|cc|r}
\toprule
 &  &
\multicolumn{8}{c|}{\textbf{Indoor}} &
\multicolumn{4}{c|}{\textbf{Outdoor}} &
\multicolumn{2}{c|}{} \textbf{Average} \\ \midrule
\multirow{2}{*}{\textbf{Type}} & \multirow{2}{*}{\textbf{Method}} &
\multicolumn{2}{c}{ScanNet} &
\multicolumn{2}{c}{IBims-1} &
\multicolumn{2}{c}{VOID} &
\multicolumn{2}{c|}{NYUv2} &
\multicolumn{2}{c}{KITTI} &
\multicolumn{2}{c|}{DDAD} &
\multicolumn{2}{c|}{} \\
 &  & MAE$\downarrow$ & RMSE$\downarrow$ &
  MAE$\downarrow$ & RMSE$\downarrow$ &
  MAE$\downarrow$ & RMSE$\downarrow$ &
  MAE$\downarrow$ & RMSE$\downarrow$ &
  MAE$\downarrow$ & RMSE$\downarrow$ &
  MAE$\downarrow$ & RMSE$\downarrow$ &
  MAE$\downarrow$ & RMSE$\downarrow$ & Rank (Count)\\
\midrule

\multirow{10}{*}{\rotatebox{90}{\textsc{Discriminative}}} & NLSPN \cite{nlspn} \textcolor{gray}{\scriptsize (ECCV '20)} &
0.036 & 0.127 &
\textbf{0.049} & 0.191 &
0.210 & 0.668 &
0.440 & 0.716 &
1.335 & 2.076 &
2.498 & 9.231 &
0.761 & 2.168 & 8.50 (12) \\

& CFormer \cite{completion-former} \textcolor{gray}{\scriptsize (CVPR '23)} &
0.120 & 0.232 &
0.058 & 0.206 &
0.216 & 0.726 &
0.186 & 0.374 &
0.952 & 1.935 &
2.518 & 9.471 &
0.675 & 2.157 & 10.00 (12) \\

& SpAgNet \cite{sparsity-agnostic-dc} \textcolor{gray}{\scriptsize (WACV '23)} & 
-- & -- & -- & -- &
0.244 & 0.706 &
0.158 & 0.292 &
0.518 & 1.788 &
4.578 & 13.236 &
-- & -- & 10.13 \; (8) \\

& BP-Net \cite{bp-net} \textcolor{gray}{\scriptsize (CVPR '24)} &
0.122 & 0.212 &
0.078 & 0.289 &
0.270 & 0.742 &
-- & -- &
-- & -- &
2.270 & 8.344 &
-- & -- & 11.75 \; (8) \\

& VPP4DC \cite{vpp4dc} \textcolor{gray}{\scriptsize (3DV '24)} &
0.023 & 0.076 &
0.062 & 0.228 &
\textbf{0.148} & \textbf{0.543} &
0.077 & 0.247 &
\textbf{0.413} & \underline{1.609} &
\textbf{1.344} & \underline{6.781} &
\textbf{0.344} & \underline{1.581} & \underline{4.00} (12) \\

& OGNI-DC \cite{ogni-dc} \textcolor{gray}{\scriptsize (ECCV '24)} &
0.029 & 0.094 &
0.059 & 0.186 &
0.175 & 0.593 &
-- & -- &
-- & -- &
\underline{1.867} & 6.876 &
-- & -- & 4.88 \; (8) \\

& DepthLab \cite{depthlab} \textcolor{gray}{\scriptsize (arXiv preprint '24)} &
0.051 & 0.081 &
0.098 & 0.198 &
0.214 & 0.602 &
0.184 & 0.276 &
0.921 & 2.171 &
4.498 & 8.379 &
0.994 & 1.951 & 8.58 (12) \\

& Prompt Depth Anything \textcolor{gray}{\scriptsize (CVPR '25)} &
0.042 & 0.079 &
0.088 & 0.196 &
0.191 & 0.605 &
0.110 & 0.233 &
0.934 & 2.803 &
2.107 & 7.494 &
0.579 & 1.902 & 6.83 (12) \\

& 
 DMD$^{3}$C \cite{liang2025distilling} \textcolor{gray}{\scriptsize (CVPR '25)} &
0.210 & 0.101 &
-- & -- &
0.225 & 0.676 &
-- & -- &
-- & -- &
2.498 & 7.766 &
-- & -- & 10.00 \;  (6)\\

& GBPN \cite{tang2026gaussian} \textcolor{gray}{\scriptsize (arXiv preprint '26)} &
-- & -- &
-- & -- &
0.220 & 0.680 &
-- & -- &
-- & -- &
-- & -- &
-- & -- & 12.50 \; (2) \\

\midrule\midrule

& \cellcolor[gray]{.97} Marigold + optim \textcolor{gray}{\scriptsize (CVPR '24)} &
\cellcolor[gray]{.97} 0.091 & \cellcolor[gray]{.97} 0.141 &
\cellcolor[gray]{.97} 0.167 & \cellcolor[gray]{.97} 0.300 &
\cellcolor[gray]{.97} 0.261 & \cellcolor[gray]{.97} 0.652 &
\cellcolor[gray]{.97} 0.194 & \cellcolor[gray]{.97} 0.309 &
\cellcolor[gray]{.97} 1.765 & \cellcolor[gray]{.97} 3.361 &
\cellcolor[gray]{.97} 22.872 & \cellcolor[gray]{.97} 32.661 &
\cellcolor[gray]{.97} 4.225 & \cellcolor[gray]{.97} 6.237 & \cellcolor[gray]{.97} 13.25 (12) \\

\multirow{6}{*}{\rotatebox{90}{\textsc{Diffusion}}} & \cellcolor[gray]{.97} Marigold + LS \cite{marigold} \textcolor{gray}{\scriptsize (CVPR '24)} &
\cellcolor[gray]{.97} 0.083 & \cellcolor[gray]{.97} 0.129 &
\cellcolor[gray]{.97} 0.154 & \cellcolor[gray]{.97} 0.286 &
\cellcolor[gray]{.97} 0.238 & \cellcolor[gray]{.97} 0.628 &
\cellcolor[gray]{.97} 0.190 & \cellcolor[gray]{.97} 0.294 &
\cellcolor[gray]{.97} 1.709 & \cellcolor[gray]{.97} 3.305 &
\cellcolor[gray]{.97} 8.217 & \cellcolor[gray]{.97} 14.728 &
\cellcolor[gray]{.97} 1.765 & \cellcolor[gray]{.97}  3.228 & \cellcolor[gray]{.97} 12.08 (12) \\

& \cellcolor[gray]{.97} Marigold-E2E + LS \cite{marigold-e2e} \textcolor{gray}{\scriptsize (WACV '25)} &
\cellcolor[gray]{.97} 0.073 & \cellcolor[gray]{.97} 0.116 &
\cellcolor[gray]{.97} 0.143 & \cellcolor[gray]{.97} 0.275 &
\cellcolor[gray]{.97} 0.233 & \cellcolor[gray]{.97} 0.623 &
\cellcolor[gray]{.97} 0.134 & \cellcolor[gray]{.97} 0.224 &
\cellcolor[gray]{.97} 1.591 & \cellcolor[gray]{.97} 3.214 &
\cellcolor[gray]{.97} 7.901 & \cellcolor[gray]{.97} 14.231 &
\cellcolor[gray]{.97} 1.679 & \cellcolor[gray]{.97} 3.114 & \cellcolor[gray]{.97} 10.42 (12) \\

\cmidrule(lr){2-17}

& \cellcolor[gray]{.97} \textcolor{gray}{Marigold-DC w/ ensemble \cite{marigold-dc}}&
\cellcolor[gray]{.97} \textcolor{gray}{0.017} &  \cellcolor[gray]{.97} \textcolor{gray}{0.057} & 
\cellcolor[gray]{.97} \textcolor{gray}{0.045} & \cellcolor[gray]{.97} \textcolor{gray}{0.166} & 
\cellcolor[gray]{.97} \textcolor{gray}{0.152} &  \cellcolor[gray]{.97} \textcolor{gray}{0.551} & 	
\cellcolor[gray]{.97} \textcolor{gray}{0.048} &  \cellcolor[gray]{.97} \textcolor{gray}{0.124} & 
\cellcolor[gray]{.97} \textcolor{gray}{0.434} & \cellcolor[gray]{.97} \textcolor{gray}{1.465} & 
\cellcolor[gray]{.97} \textcolor{gray}{2.364} & \cellcolor[gray]{.97} \textcolor{gray}{6.449} &
\cellcolor[gray]{.97} \textcolor{gray}{0.510} & \cellcolor[gray]{.97} \textcolor{gray}{1.469} & \cellcolor[gray]{.97} \textcolor{gray}{1.75 (12)}
\\

\cmidrule(lr){2-17}

 & \cellcolor[gray]{.9} \textbf{Marigold-DC \cite{marigold-dc} \textcolor{gray}{\scriptsize (ICCV '25)}\textbf} &
\cellcolor[gray]{.9} \textbf{0.020} & \cellcolor[gray]{.9}  \underline{0.063} &
\cellcolor[gray]{.9}  0.062 & \cellcolor[gray]{.9}  0.205 &
\cellcolor[gray]{.9}  \underline{0.157} & \cellcolor[gray]{.9}  \underline{0.557} &
\cellcolor[gray]{.9}  0.057 & \cellcolor[gray]{.9}  0.142 &
\cellcolor[gray]{.9}  0.558 & \cellcolor[gray]{.9}  1.676 &
\cellcolor[gray]{.9}  2.985 & \cellcolor[gray]{.9}  7.905 &
\cellcolor[gray]{.9}  0.640 & \cellcolor[gray]{.9}  1.758 & \cellcolor[gray]{.9} 5.08 (12) \\

& \cellcolor[gray]{.9}  \textbf{Marigold-SSD\APLstar} \textcolor{gray}{(Ours)} &
\cellcolor[gray]{.9}  \underline{0.022} & \cellcolor[gray]{.9}  \textbf{0.062} &
\cellcolor[gray]{.9}  \underline{0.056} & \cellcolor[gray]{.9}  \textbf{0.182} &
\cellcolor[gray]{.9}  0.177 & \cellcolor[gray]{.9}  0.588 &
\cellcolor[gray]{.9}  \textbf{0.045} & \cellcolor[gray]{.9}  \textbf{0.128} &
\cellcolor[gray]{.9}  2.443 & \cellcolor[gray]{.9}  4.070 &
\cellcolor[gray]{.9}  3.855 & \cellcolor[gray]{.9}  7.840 &
\cellcolor[gray]{.9}  1.100 & \cellcolor[gray]{.9}  2.145  & \cellcolor[gray]{.9} 5.33 (12) \\

& \cellcolor[gray]{.9}  \textbf{Marigold-SSD \textcolor{gray}{(Ours)}} &
\cellcolor[gray]{.9}  0.027 & \cellcolor[gray]{.9}  0.068 &
\cellcolor[gray]{.9}  0.060 & \cellcolor[gray]{.9}  \underline{0.185} &
\cellcolor[gray]{.9}  0.182 & \cellcolor[gray]{.9}  0.590 &
\cellcolor[gray]{.9}  \underline{0.052} & \cellcolor[gray]{.9}  \underline{0.134} &
\cellcolor[gray]{.9}  \underline{0.454} & \cellcolor[gray]{.9}  \textbf{1.496} &
\cellcolor[gray]{.9}  2.065 & \cellcolor[gray]{.9}  \textbf{6.522} &
\cellcolor[gray]{.9}  \underline{0.473} & \cellcolor[gray]{.9}  \textbf{1.499} & \cellcolor[gray]{.9}  \textbf{3.75} (12) \\
\bottomrule
\end{tabular}}
\end{table*}

%% file: figures/qualitative/qualitative.tex
\begin{figure}[!htb]
\centering
    \centering
    \includegraphics[width=1.0\linewidth]{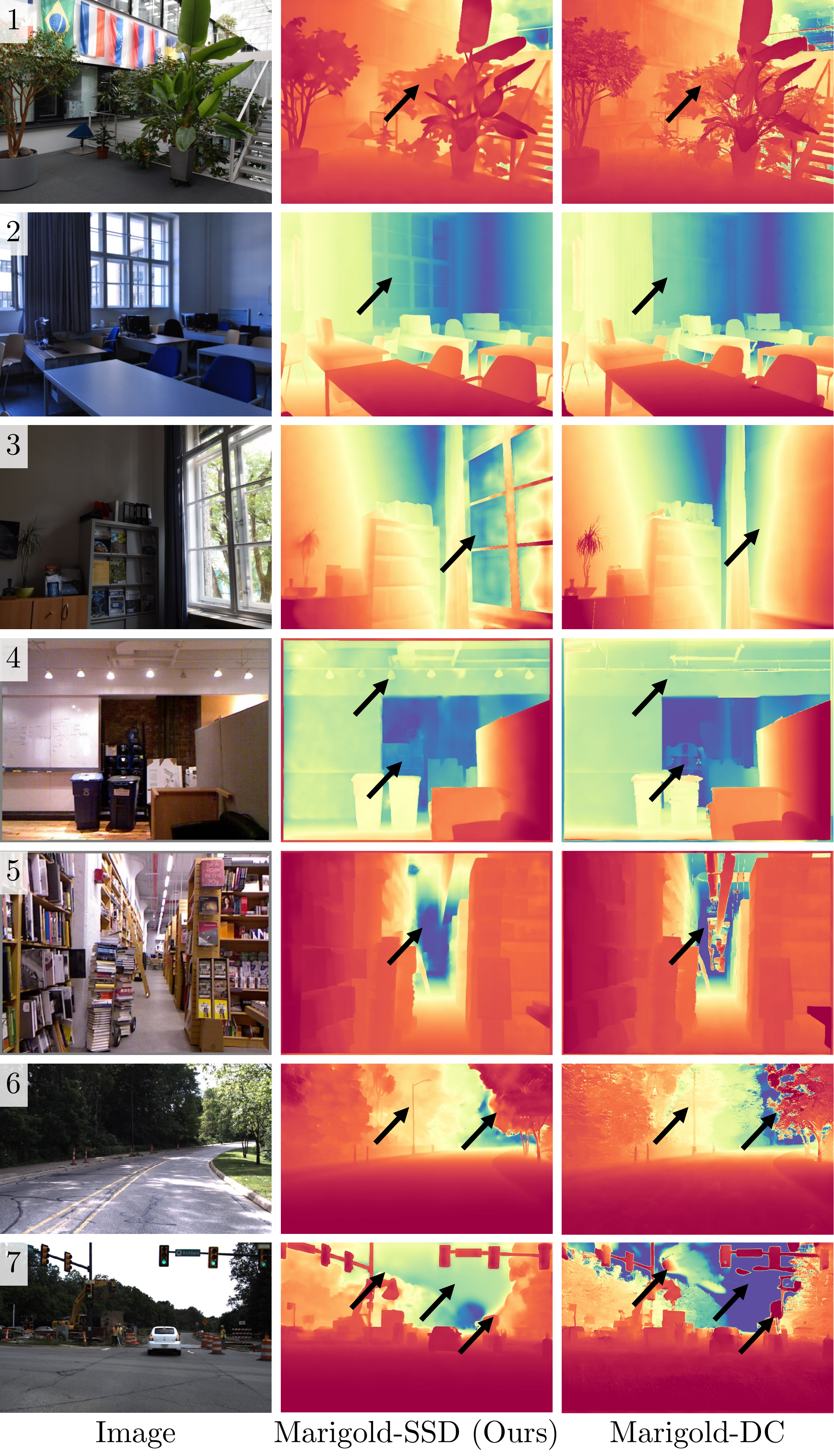}
    \vspace{-0.28cm}
    \caption{
    \textbf{Qualitative results.} Marigold-SSD generally produces smoother depth maps 
    than Marigold-DC~\cite{marigold-dc}, which tends to over-refine details that can 
    lead to unrealistic scene structures. The black arrows highlight variations in the 
    estimated depth, while the red and blue colors indicate the nearest and farthest 
    regions.}
    \vspace{-1cm}
    \label{fig:qualitative}
\end{figure}

%% file: figures/qualitative/qualitative-kitti.tex
\begin{figure*}[ht]
\centering
    \centering
    \includegraphics[width=1.0\textwidth]{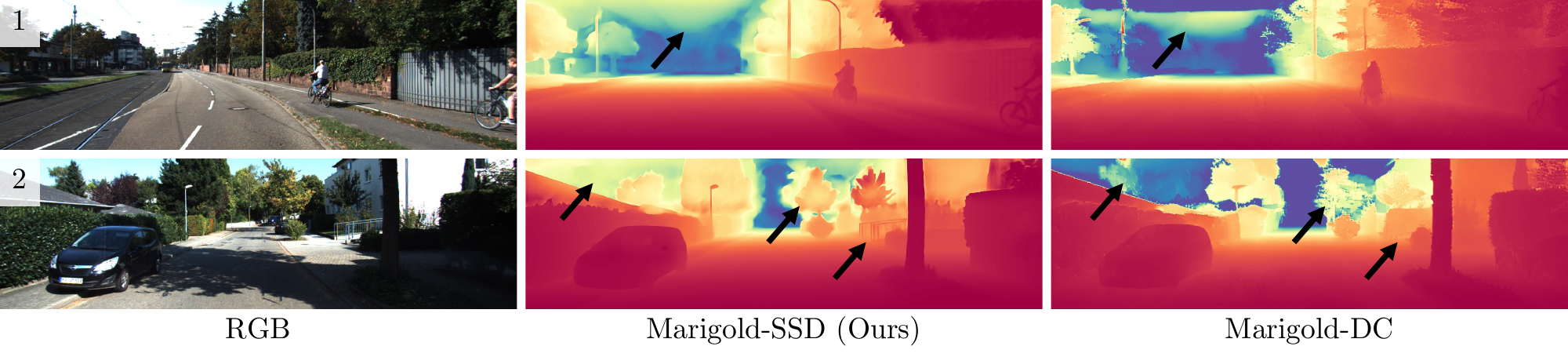}
    \caption{\textbf{Qualitative results.} Both Marigold-SSD and Marigold-DC tend to underestimate sky depth on KITTI and DDAD, consistent with prior Marigold limitations and limited conditioning information in the sky, while they differ in how they estimate fine scene details.}
    \label{fig:qualitative-kitti}
\end{figure*}

%% file: figures/sparsity-levels/sparsity-levels.tex
\begin{figure}[ht]
    \hspace{0.1px} 
	\subfloat{\includegraphics[width = 0.49\linewidth]{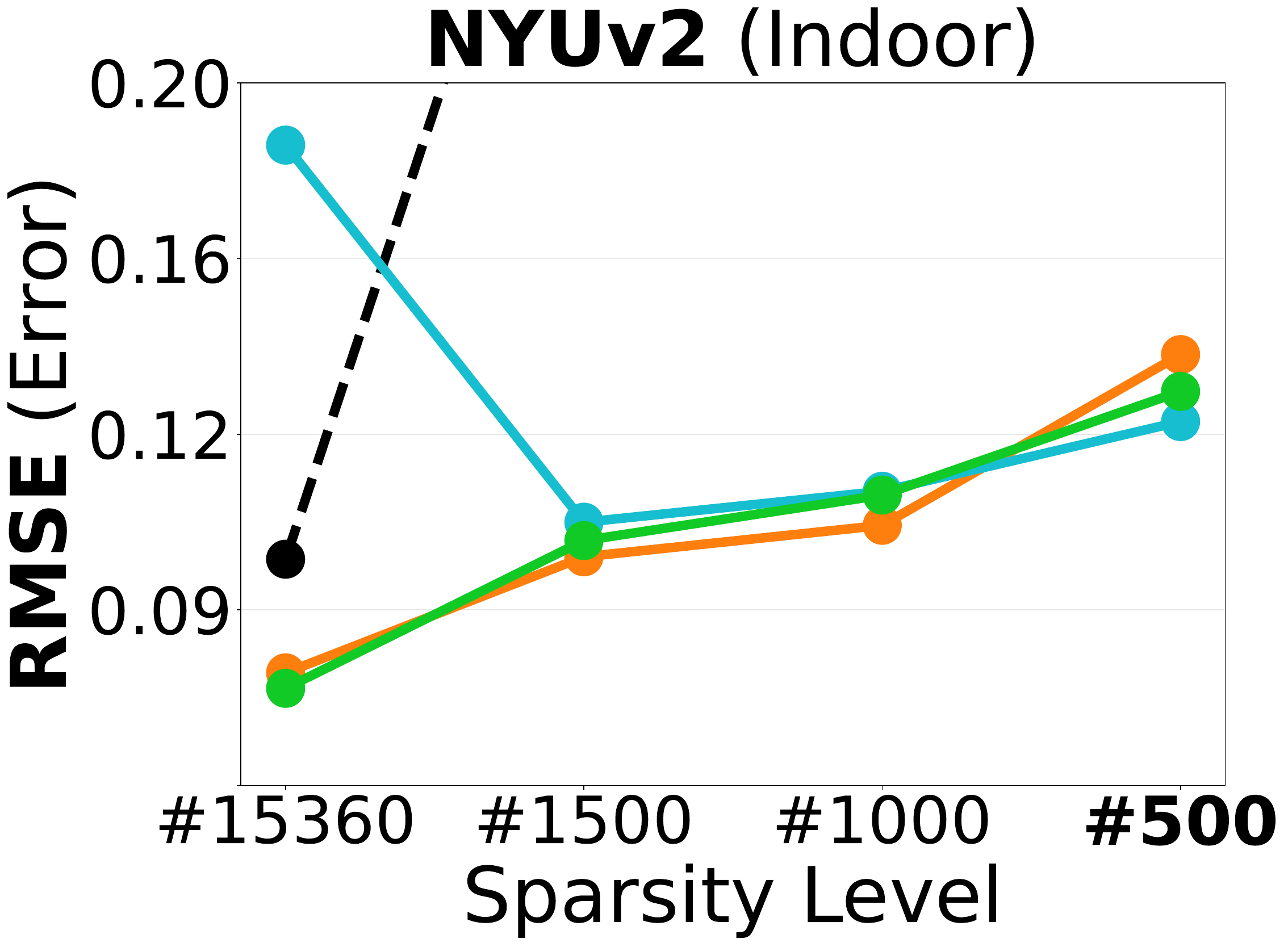}}
    \hspace{0.1px} 
	\subfloat{\includegraphics[width = 0.49\linewidth]{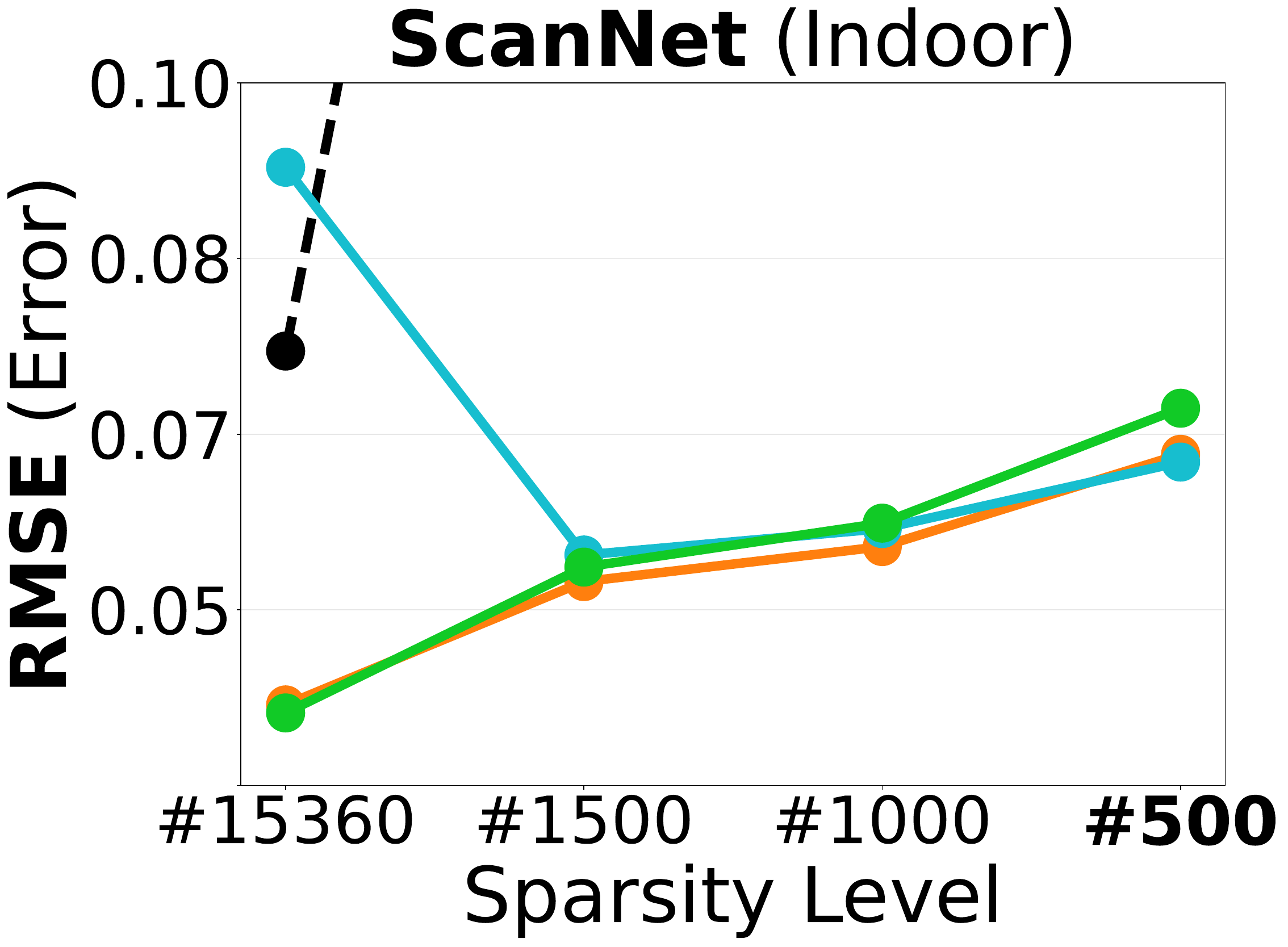}} \\
    \hspace{0.1px}
    \subfloat{\includegraphics[width = 1.0\linewidth]{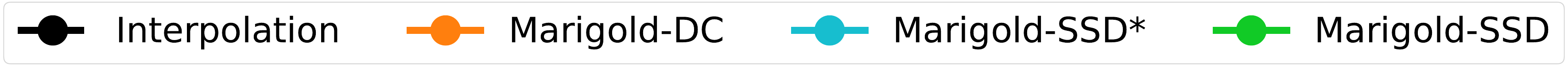}}
	\caption{
    \textbf{Evaluation under multiple levels of depth density} on NYUv2 
    and ScanNet. Depth density is denoted by the number of depth samples (\#). 
    See the supplementary material for all datasets.
    }
    \label{fig:sparsity-levels}
\end{figure}

%% file: figures/sparsity-levels/sparsity-levels-ddad.tex
\begin{figure}[h]
	\subfloat{\includegraphics[width = 0.49\linewidth]{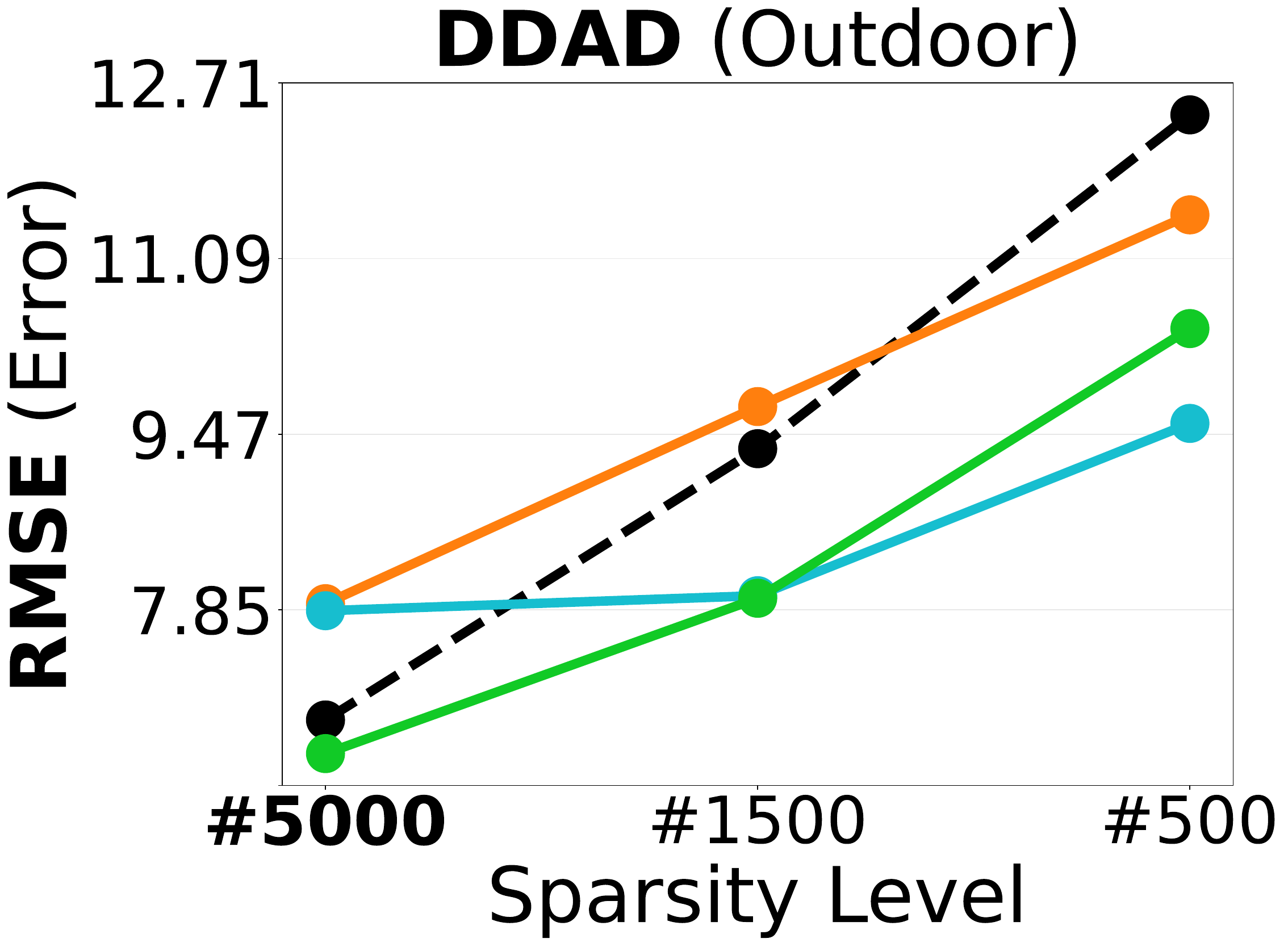}}
    \hspace{0.1px}
    \subfloat{\includegraphics[width = 0.49\linewidth]{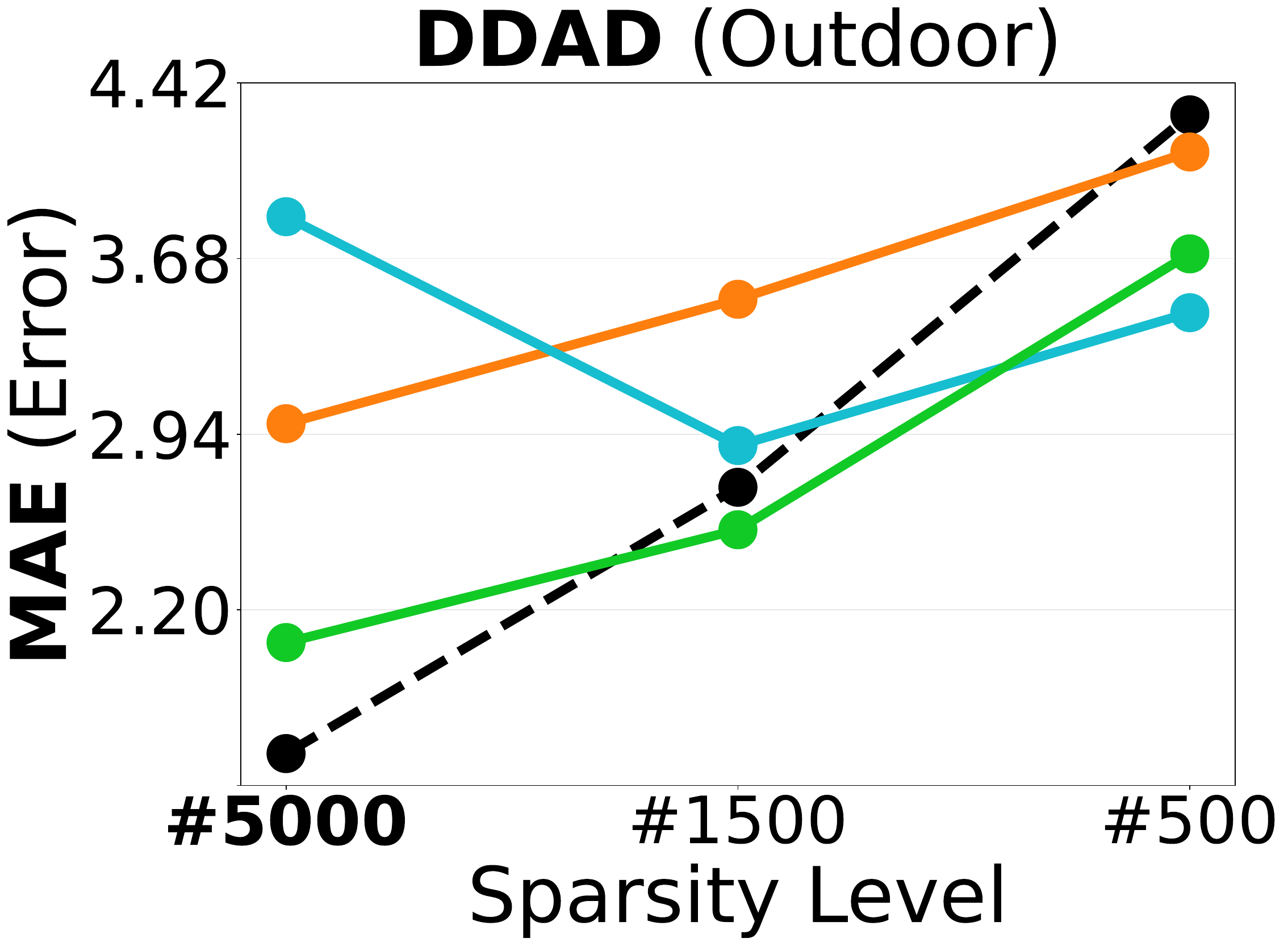}} \\
    \hspace{0.1px}
    \subfloat{\includegraphics[width = 1.0\linewidth]{figures/sparsity-levels/legend-horizontal.pdf}}
    \caption{
    \textbf{Challenging the models on DDAD.}
    At the commonly used sparsity level of 5000 points even sophisticated models
    can be outperformed by trivial Barycentric interpolation. 
    }
    \label{fig:sparsity-levels-ddad}
\end{figure}

%% file: 5_results.tex
\subsection{Runtime Analysis}
We evaluate the efficiency of our method and report relative speedup over Marigold-DC~\cite{marigold-dc} 
in Tab.~\ref{tab:timing}. All timings are measured on an NVIDIA RTX~4090 GPU and Marigold-DC is timed 
without ensembling (single run). Our method achieves an average $66\times$ speedup across indoor and 
outdoor datasets while also achieving better average performance (Tab.~\ref{tab:quantitative-results}). 
Running the standard 10-sample ensembling strategy in Marigold-DC increases runtime approximetaly linearly 
and would result in \emph{$660\times$ speedup} on average. Furthermore, we analyze the speed--performance 
tradeoff on KITTI~\cite{kitti-dataset} in Tab.~\ref{tab:timing-kitti} and Fig.~\ref{fig:compare-speed} by 
comparing diffusion-based and discriminative depth completion methods. Marigold-SSD substantially narrows 
the efficiency gap to discriminative approaches retaining the benefits of a strong diffusion prior with 
runtime comparable to the discriminative models.

\subsection{Zero-Shot Depth Completion Results}
We evaluate our models in zero-shot context and present the results in Tab.~\ref{tab:quantitative-results}. 
We compare with discriminative methods~\cite{nlspn,completion-former,sparsity-agnostic-dc,bp-net,vpp4dc,
ogni-dc,depthlab,prompt-da,DMD3C,tang2026gaussian} and diffusion-based methods~\cite{marigold,marigold-e2e,marigold-dc}.
For depth completion methods~\cite{marigold,marigold-e2e} the sparse depth was used only for
aligning shift and scale of the result using: least squares (denoted by LS) or L1 + L2 optimization 
(denoted by "optim").
We achieve the best average \textbf{RMSE of 1.499} and  \textbf{MAE of 0.473} versus 1.758 and 0.640 
respectively for Marigold-DC  without ensembling. When ensembling is utilized, Marigold-DC achieves 
a RMSE of 1.469 and a MAE of 0.510 at the expense of an order of magnitude slower inference. Considering 
the \textit{need for speed}, we run and compare with Marigold-DC without ensembling in the rest of the paper. 
Qualitative results on several datasets of Marigold-SSD compared to Marigold-DC can be seen in 
Fig.~\ref{fig:qualitative}~\&~\ref{fig:qualitative-kitti}.

\subsection{Evaluation under Varying Depth Sparsity}
We evaluate our method across a range of sparsity levels, see Fig.~\ref{fig:sparsity-levels}~\&~\ref{fig:sparsity-levels-ddad}. 
Additionally, we assess a Barycentric interpolation computed within 
Delaunay triangulation of the sparse depth condition. The interpolation disregards 
RGB images and sets depth values outside the convex hull to zero. Visualizations of the 
sparse and interpolated depth are provided in the supplementary material.
As expected, performance improves with denser conditioning depth. When the density 
reaches about 5\% (15360 points), interpolation can achieve competitive performance.
On DDAD, at 5000 points, interpolation appears to outperform Marigold-DC and Marigold-SSD. 
At lower densities Marigold-SSD outperforms both Marigold-DC and simple interpolation.

\subsection{Ablation Studies}

In this section, we compare our late-fusion strategy against several early-fusion alternatives, and provide an ablation study on the sampling condition density during fine-tuning.

\noindent \textbf{Late vs Early fusion.} We consider two early-fusion approaches: i) \textit{Frozen VAE} and ii) \textit{Conditional Encoder}. \textit{Frozen VAE} encodes the depth condition using the frozen encoder $\mathcal{E}$ and passes it to the UNet through extra input channels, following the RGB conditioning of Marigold~\cite{marigold}. We evaluate this variant under two types of conditioning: sparse depth and pre-completed depth. Pre-completion is performed using the same barycentric interpolation tested earlier under varying sparsity. During fine-tuning, only the UNet is updated with a learning rate of $3 \times 10^{-5}$. \textit{Conditional Encoder} is an early-fusion counterpart to our conditional decoder, where the depth condition is encoded by a duplicated branch of the VAE encoder $\mathcal{E}$. The fine-tuning follows the same protocol in our late fusion strategy, updating the UNet and the conditional encoder, while the VAE decoder $\mathcal{D}$ stays frozen. Results in Tab.~\ref{tab:ablations-early-fusion} show that all early-fusion variants perform worse on average than our late-fusion approach. 

\input{tables/ablations-early-fusion}

\noindent \textbf{Sampling Condition Density.} To study the effect of condition sampling density during fine-tuning and performance under out-of-distribution sparsity, we fine-tune 3 additional versions of our model using different sampling densities: (A) constant density $0.16\%$ ($\sim500$ points), (B) densities in $\left[0.16\%, 0.32\%\right]$ ($\left[\sim500, \sim1000\right]$ points), and (C) constant density $0.5\%$ ($\sim1500$ points). Point counts correspond to a resolution of $640\times480$. Results for ScanNet and DDAD are shown in Fig.~\ref{fig:training-sparsity-ablations}.  Zero-shot performance degrades under narrower training sparsity regimes, supporting our default choice of fine-tuning over a broader range. Results for the remaining evaluation datasets are provided in the supplementary material.

\input{figures/training-sparsity-ablation/training-sparsity-ablation}

%% file: tables/ablations-early-fusion.tex
\begin{table*}[t]
\caption{
\textbf{Late vs Early Fusion.} Comparison with early-fusion:  models with (i) frozen VAE or (ii) a trainable conditional encoder. i.e, an early fusion equivalent of our conditional decoder. Models trained with lower density-level are denoted by \APLstar. The \textbf{best} and \underline{second best} results are highlighted for each training setup. Our late fusion strategy outperforms in almost all cases early-fusion approaches.
}
\label{tab:ablations-early-fusion}
\centering
\resizebox{\textwidth}{!}{%
\begin{tabular}{lccccccccccccc}
\toprule
\multirow{2}{*}{Method}     &
\multirow{2}{*}{Condition}  &
\multicolumn{2}{c}{ScanNet} &
\multicolumn{2}{c}{IBims-1} &
\multicolumn{2}{c}{VOID}    &
\multicolumn{2}{c}{NYUv2}   &
\multicolumn{2}{c}{KITTI}   &
\multicolumn{2}{c}{DDAD}    \\
&
&
MAE$\downarrow$ & RMSE$\downarrow$ &
MAE$\downarrow$ & RMSE$\downarrow$ &
MAE$\downarrow$ & RMSE$\downarrow$ &
MAE$\downarrow$ & RMSE$\downarrow$ &
MAE$\downarrow$ & RMSE$\downarrow$ &
MAE$\downarrow$ & RMSE$\downarrow$ \\
\midrule

Frozen VAE\APLstar &
\textbf{Sparse} &
\underline{0.038} & 0.075             &
0.099             & 0.212             &
0.254             & 0.721             &
0.089             & \underline{0.173} &
1.899             & 3.642             &
6.612             & 12.295            \\

Frozen VAE\APLstar &
\textbf{Interpolated} &
\textbf{0.022}    & \textbf{0.062}    &
\underline{0.057} & \textbf{0.178}    &
\underline{0.208} & 0.663             &
\underline{0.064} & 0.185             &
\textbf{0.978}    & \textbf{2.418}    &
\underline{3.897} & \underline{9.473} \\

Conditional Encoder\APLstar &
\textbf{Sparse} &
0.063             & 0.103             &
0.131             & 0.257             &
0.222             & \underline{0.615} &
0.124             & 0.208             &
\underline{1.521} & \underline{3.059} &
7.637             & 13.779 \\

\cellcolor[gray]{.9}  \textbf{Marigold-SSD\APLstar} &
\cellcolor[gray]{.9}  \textbf{Sparse} &
\cellcolor[gray]{.9}  \textbf{0.022}  & \cellcolor[gray]{.9}  \textbf{0.062}    &
\cellcolor[gray]{.9}  \textbf{0.056}  & \cellcolor[gray]{.9}  \underline{0.182} &
\cellcolor[gray]{.9}  \textbf{0.177}  & \cellcolor[gray]{.9}  \textbf{0.588}    &
\cellcolor[gray]{.9}  \textbf{0.045}  & \cellcolor[gray]{.9}  \textbf{0.128}    &
\cellcolor[gray]{.9}  2.443           & \cellcolor[gray]{.9}  4.070             &
\cellcolor[gray]{.9}  \textbf{3.855}  & \cellcolor[gray]{.9}  \textbf{7.840}    \\

\midrule
\midrule

Frozen VAE &
\textbf{Sparse} &
0.045 & 0.087  &
0.111 & 0.237  &
0.315 & 0.820  &
0.105 & 0.201  &
1.280 & 2.860  &
5.976 & 11.511 \\

Frozen VAE &
\textbf{Interpolated} &
\textbf{0.024}    & \underline{0.070} &
\textbf{0.054}    & \textbf{0.175}    &
0.224             & 0.711             &
\underline{0.064} & 0.220             &
\underline{0.559} & \underline{1.791} &
\underline{2.223} & \underline{7.009} \\

Conditional Encoder &
\textbf{Sparse} &
0.051             & 0.088              &
0.112             & 0.232              &
\underline{0.216} & \underline{0.619}  &
0.106             & \underline{0.185}  &
1.372             & 2.734              &
6.615             & 12.152             \\

\cellcolor[gray]{.9}  \textbf{Marigold-SSD} &
\cellcolor[gray]{.9}  \textbf{Sparse} &
\cellcolor[gray]{.9}  \underline{0.027} & \cellcolor[gray]{.9}  \textbf{0.068}    &
\cellcolor[gray]{.9}  \underline{0.060} & \cellcolor[gray]{.9}  \underline{0.185} &
\cellcolor[gray]{.9}  \textbf{0.182}    & \cellcolor[gray]{.9}  \textbf{0.590}    &
\cellcolor[gray]{.9}  \textbf{0.052}    & \cellcolor[gray]{.9}  \textbf{0.134}    &
\cellcolor[gray]{.9}  \textbf{0.454}    & \cellcolor[gray]{.9}  \textbf{1.496} &
\cellcolor[gray]{.9}  \textbf{2.065}    & \cellcolor[gray]{.9}  \textbf{6.522} \\

\bottomrule
\end{tabular}}
\end{table*}

%% file: figures/training-sparsity-ablation/training-sparsity-ablation.tex
\begin{figure}[h]
	\subfloat{\includegraphics[width = 0.49\linewidth]{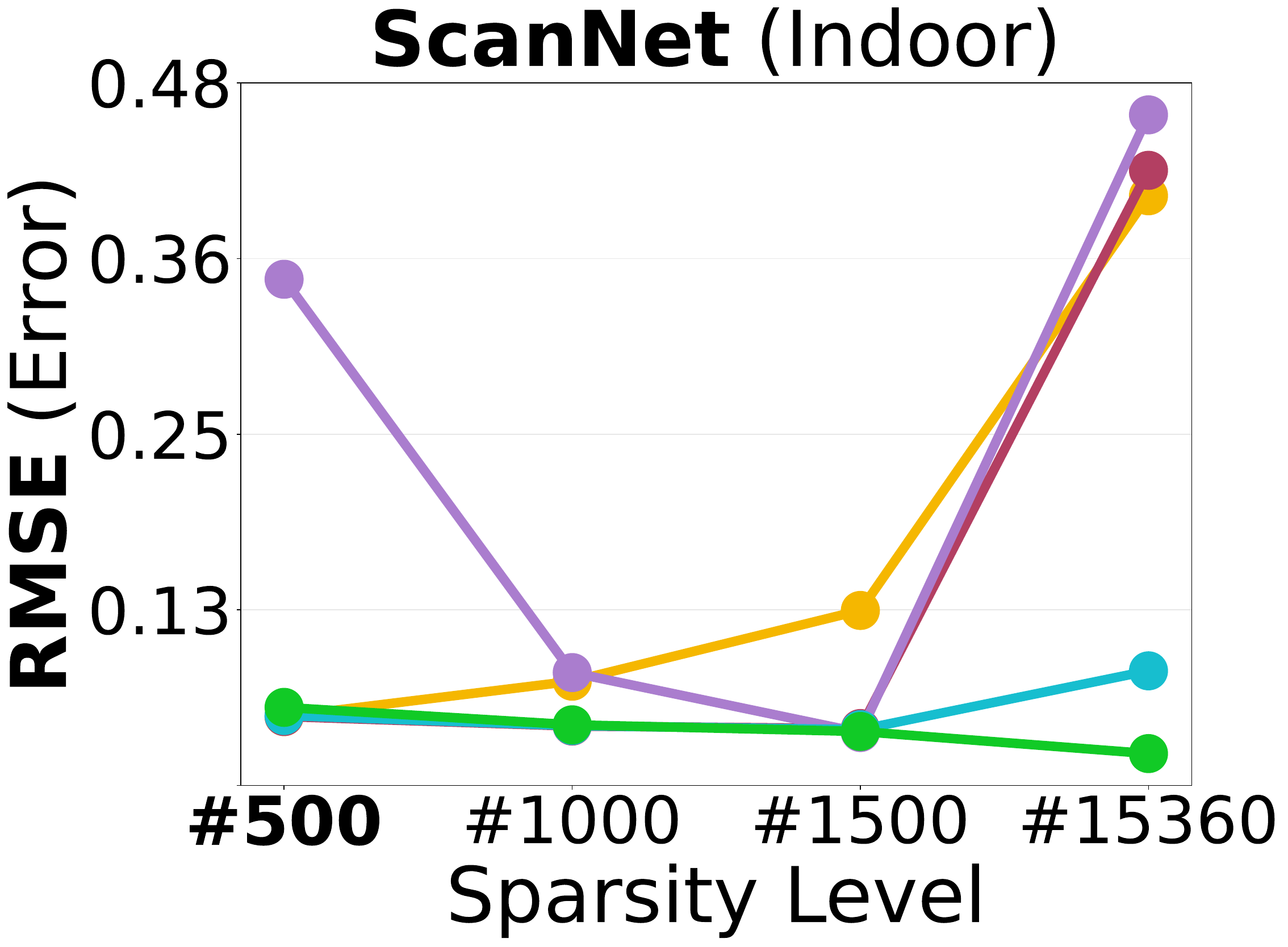}}
    \hspace{0.1px}
    \subfloat{\includegraphics[width = 0.49\linewidth]{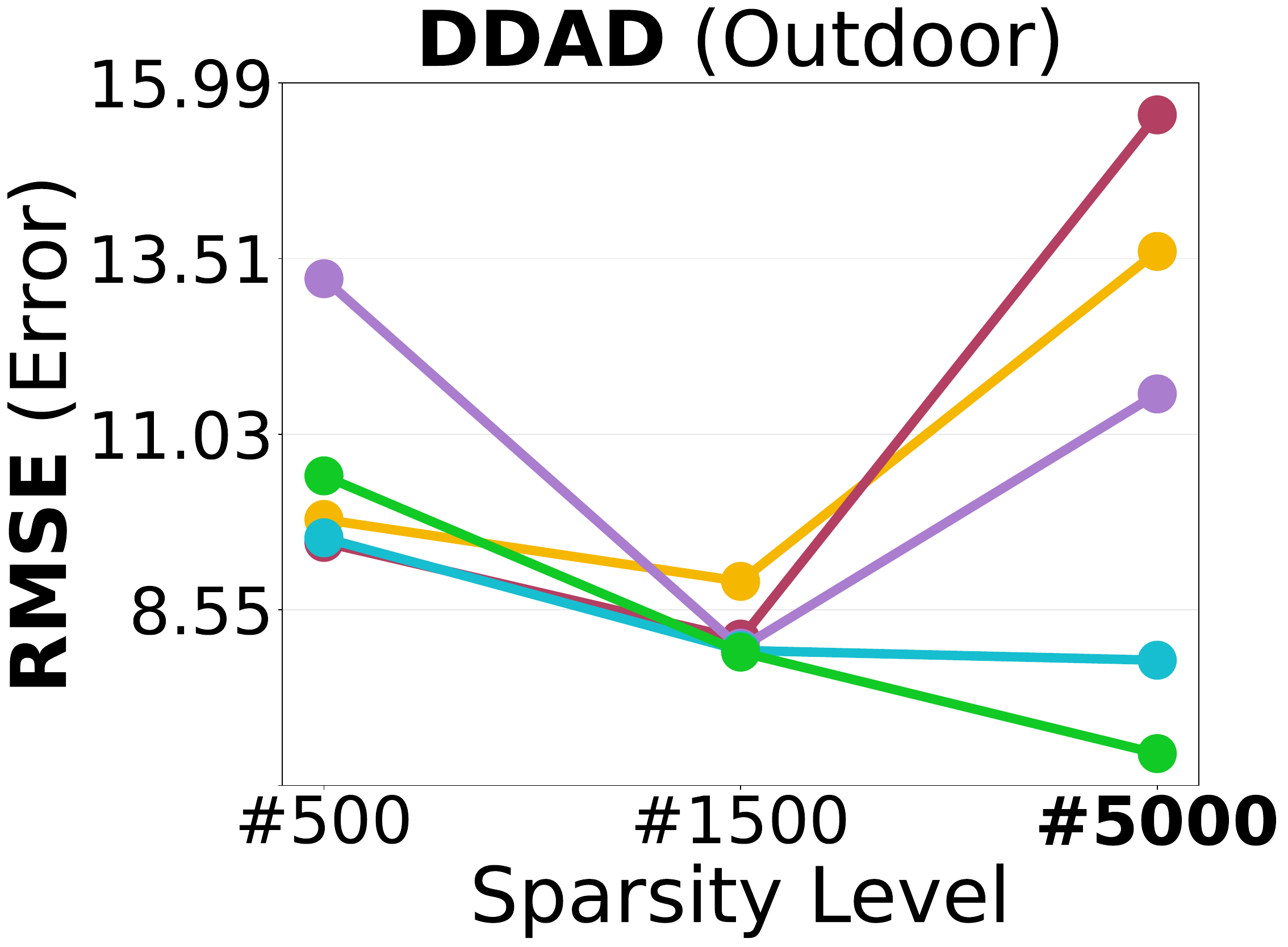}} \\
    \subfloat{\includegraphics[width = 1.00\linewidth]{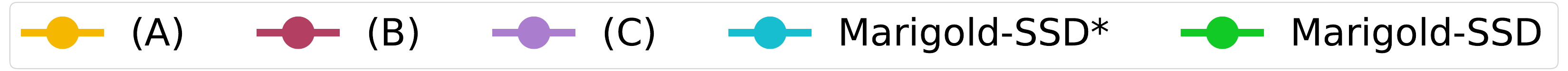}} 
    \caption{
    \textbf{Sampling Density.} 
    Models (A), (B) \& (C) fine-tuned on different densities. See the supplementary material for all datasets.
    }
    \label{fig:training-sparsity-ablations}
\end{figure}

%% file: 6_discussion.tex
\section{Discussion}
\noindent\textbf{Preserving Diffusion Prior.} Our results show that the proposed architecture 
and fine-tuning framework effectively leverage Marigold’s pre-trained knowledge for zero-shot 
depth completion. Marigold-SSD performs better than Marigold-DC on all datasets except VOID. 
While ensembling improves Marigold-DC, it requires an order of magnitude more computation, 
and our method remains competitive while being $660\times$ faster. Marigold-SSD substantially 
narrows the efficiency gap to discriminative approaches while retaining the benefits of a strong 
diffusion prior.
Compared to the strongest discriminative baseline, VPP4DC~\cite{vpp4dc}, Marigold-SSD requires 
fine-tuning only a single model on synthetic datasets. In contrast, the VPP4DC results 
in Tab.~\ref{tab:quantitative-results} are obtained from 3 separately trained 
models~\cite{vpp4dc,marigold-dc}: one trained on SceneFlow~\cite{sceneflow-dataset}, 
one on SceneFlow + KITTI, and one on SceneFlow + NYUv2. This makes its deployment zero-shot 
generalization less clear, making Marigold-SSD more compelling for general applications.
Ablations on early-fusion reveal, that the off-the-shelf VAE encoder is ill-suited for 
sparse inputs and refining or rethinking sparse-data fusion is necessary. Although 
pre-completion helps to increase the performance, our late-fusion approach performs 
better than early-fusion strategies. As illustrated in Fig.~\ref{fig:qualitative}, 
our method produces smoother outputs typical in single‑step diffusion~\cite{marigold-e2e} 
or expected in aggregated ensembles of outputs, while Marigold-DC (w/o ensembling) tends 
to over-refine details. Marigold-SSD achieves high quality outputs in a single step (rows $1-3,6$), 
reducing high frequency details that can lead to unrealistic structures (rows $5,7$). 
In the supplementary material we provide quantitative evaluation of depth boundary accuracy 
and include additional qualitative examples showcasing differences in detail generation by 
the diffusion-based methods.

\noindent\textbf{Limitations.}
Our end-to-end fine-tuning requires to set sampling density range of the condition. 
As demonstrated in ablation studies, completion of out-of-distribution depth maps may 
exhibit a steep performance drop. However, Marigold‑SSD trained on our default broad 
spectrum of sparsity levels can achieve strong zero-shot performance across domains. 
The impact of condition sampling density could be less pronounced at higher densities, 
where much of the target information is already provided and lightweight Barycentric 
interpolation can achieve strong results. 
A potential future direction lies in adaptive normalization strategies and scaling depth 
condition to the VAE's operational range. This could mitigate possible depth inaccuracies, 
which may occur when depth range of the scene deviates from the range of the depth condition. 
Additionally, developing strategies to better capture outdoor depth distributions may reduce 
the characteristic bias of Marigold-based methods in sky regions (see row 7 of 
Fig.~\ref{fig:qualitative} and Fig.~\ref{fig:qualitative-kitti}).
Similar to~\cite{depth-anything}, a semantic segmentation model could assign infinite depth 
to sky regions during training. Alternatively, as in~\cite{depth-anything-v3}, the model 
could be trained to predict sky masks which may help to avoid degraded predictions in 
regions where ground-truth is unavailable. \\

\noindent\textbf{Performance under Variety of Sparsity Levels.}
To illustrate when depth completion models provide real value, we include a comparison 
against a simple interpolation baseline. Increasing the condition sparsity allows interpolation 
to approach performance of state‑of‑the‑art methods. The commonly used evaluation density for 
the DDAD dataset~\cite{vpp4dc, ogni-dc, marigold-dc} lies well beyond this threshold. 
Under this setting, interpolation achieves MAE 1.598 and RMSE 6.831 outperforming many 
methods in Tab.~\ref{tab:quantitative-results}.
However, the benefit of models with strong pretrained priors become clear at lower density 
levels, such as 500 points, consistent with established protocols for NYUv2 and ScanNet. 
In this regime, our method outperforms Marigold-DC. \\

%% file: 7_conclusion.tex
\noindent \textbf{Conclusion.} We presented Marigold‑SSD, a late-fusion depth‑completion 
framework achieving a $66\times$ speed‑up and improved performance compared to Marigold-DC 
(w/o ensembling). We also identified that simple interpolation can achieve competitive 
zero-shot results on DDAD dataset under standard density levels. \\

\noindent \textbf{Acknowledgments.} 
This project was supported by the EU Horizon Europe project “RoBétArmé” (Grant Agreement 101058731). 
Part of this work was conducted at ETH Zürich. HPC resources were provided by the Pioneer Centre for 
AI and DTU Computing Center \cite{DTU_DCC_resource}.